\pdfoutput=1

\documentclass[11pt]{article}

\usepackage[final]{acl}
\usepackage{booktabs} 
\usepackage{times}
\usepackage{latexsym}

\usepackage[T1]{fontenc}

\usepackage[utf8]{inputenc}
\usepackage{csquotes}
\usepackage{microtype}

\usepackage{inconsolata}

\usepackage{graphicx}
\usepackage{multirow}
\usepackage{multicol}
\title{NIM: Neuro-symbolic Ideographic Metalanguage for Inclusive Communication}

\author{
  Prawaal Sharma \\
  Infosys \\
  Pune, Maharashtra, India \\  
  \texttt{prawaal\_sharma@infosys.com}
  \And
  Poonam Goyal \\
  BITS Pilani \\
  Pilani, Rajasthan, India \\  
  \texttt{poonam@pilani.bits-pilani.ac.in}
  \AND
  Navneet Goyal \\
  BITS Pilani \\
  Pilani, Rajasthan, India \\  
  \texttt{goel@pilani.bits-pilani.ac.in}
  \And
  Vidisha Sharma \\
  BITS Pilani \\
  Goa, India \\  
  \texttt{vidishasharma@gmail.com}
}

\begin{document}
\maketitle
\begin{abstract}
Digital communication has become the cornerstone of modern interaction, enabling rapid, accessible, and interactive exchanges. However, individuals with lower academic literacy often face significant barriers, exacerbating the \enquote{digital divide}. In this work, we introduce a novel, universal ideographic metalanguage designed as an innovative communication framework that transcends academic, linguistic, and cultural boundaries. Our approach leverages principles of Neuro-symbolic AI, combining neural-based large language models (LLMs) enriched with world knowledge and symbolic knowledge heuristics grounded in the linguistic theory of Natural Semantic Metalanguage (NSM). This enables the semantic decomposition of complex ideas into simpler, atomic concepts. Adopting a human-centric, collaborative methodology, we engaged over 200 semi-literate participants in defining the problem, selecting ideographs, and validating the system. With over 80\% semantic comprehensibility, an accessible learning curve, and universal adaptability, our system effectively serves underprivileged populations with limited formal education. 
\end{abstract}

\section{Introduction}

Communication, in its many forms, has always been the bedrock of human progress, shaping societies and driving innovation. In the modern interconnected world, digital communication has emerged as a pivotal force, revolutionizing how we communicate. We propose an ideographic communication metalanguage for semi-literates, promoting socio-digital inclusion, and a more interconnected society. We define semi-literates as individuals with limited formal education typically only up to the primary or early secondary level who face significant challenges in navigating digital communication platforms due to constrained reading, writing, and digital literacy skills.

Visual communication is intuitive, easy to comprehend, language agnostic, containing spatial and directional attributes \cite{iconicinterfacing, van, review, iconji}. While this can help in the facilitation of human-computer communication specially for people with limited academic education, however having a set of visual ideographs only lacks morphology and syntax, which is ingrained in orthographic forms of communication thus making pure visual ideographic set prone to misinterpretation. Earlier studies on the use of visual communication for academically challenged individuals observe challenges with learnability (due to large inventory of ideographs), extensibility (non-universal design) along with ambiguity (lack of grammar) \cite{semantographics}.   

\begin {figure*} [t]
\centering
{\includegraphics [width=0.75\textwidth] {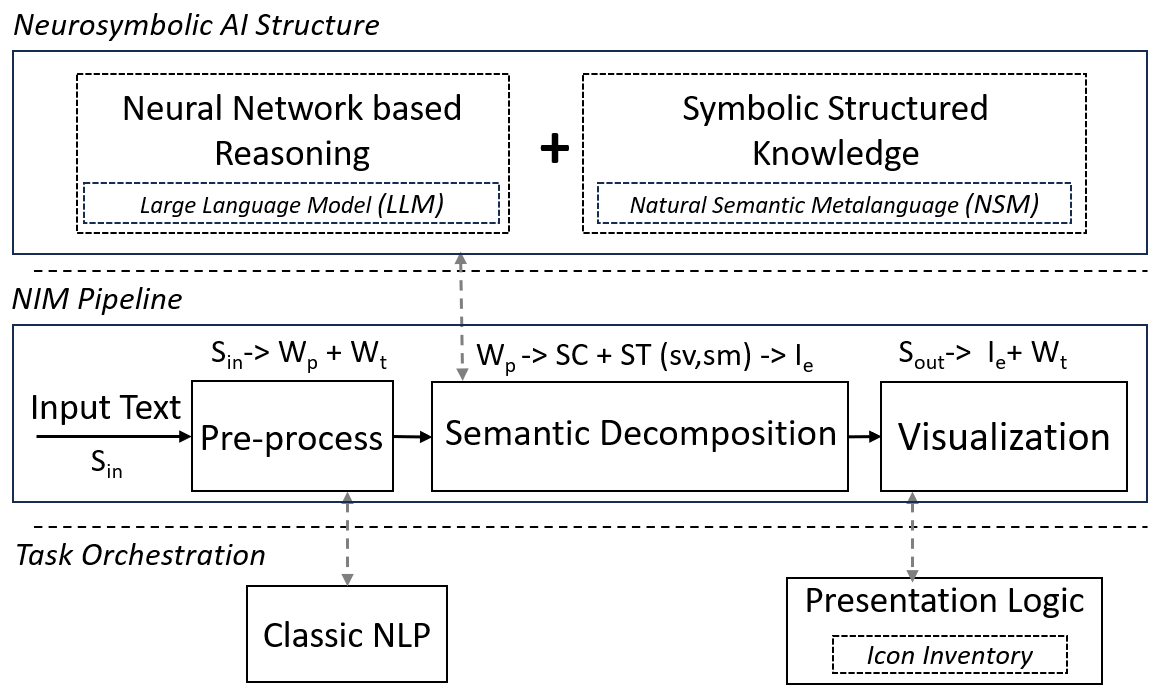}}
\caption{High Level NIM architecture, comprising of Neuro-symbolic AI, Pipeline and Orchestration layers.}
\label{summary}
\centering
\end {figure*}

Figure \ref{summary} presents the conceptual foundation of our work, embodying the neuro-symbolic AI paradigm through a synthesis of symbolic reasoning and neural parametric learning. By augmenting and combining the strength of Large Language Models (LLMs) enabling generation of new concepts, along with structured ontology as guiding principles for semantic decomposition of complex ideas into simpler and atomic concepts, we design a novel ideographic metalanguage for semi-literates. The pipeline goes through processing of input sentence $S_{in}$ into distinct parts of picturable ($W_p$) and textual (non picturable, $W_t$) sections. While $W_p$ gets broken down into ideographic elementary components ($I_e$) which comprises of Semantic Classes, Templates, Variables and Molecules ($SC/ ST/ (sv,sm)$), the non picturable sections are preserved as plain text coming together in an interactive user interface customized for semi-literate users. Consequently, we call our method as NIM (Neuro-symbolic Ideographic Metalanguage). 

NIM leverages the linguistic theory of Natural Semantic Metalanguage (NSM) \cite{nsm1, nsm2} for symbolic reasoning module achieving semantic simplification of complex concepts into more elementary elements. This symbolic reasoning is integrated with Generative Artificial Intelligence (Gen-AI) and construct the design of a visual metalanguage (including bare-minimum text, we refer as \textit{binding text}) for effective communication in semi-literate settings. We further enhance our work by making the binding text multilingual thereby making it universal for use. 

We adopt a collaborative Human-Centered Design (HCD) methodology that actively involves semi-literate individuals across the development cycle, facilitating authentic problem discovery, participatory design, and rigorous empirical validation. The validation spans multiple dimensions, including comprehensibility, user engagement, ideographic effectiveness, and extensibility. Beyond semi-literate populations, our approach has potential applicability to diverse user groups such as individuals with intellectual disabilities, multilingual teams, and children with dyslexia.  

In summary our contribution, is three-fold, (i) We build a novel metalanguage for semi-literates, enabling them for digital communication; (ii) Our work facilitates seamless integration with digital ecosystem, for easy and intuitive human-computer interaction; and (iii) Our work is universal, extensible for multiple domains, geographies and cultures.

It can be argued that the speech-based systems and vision-language models (VLMs) offer promising modalities for accessibility and can be used to solve the problem of digital communication for semi-literate groups. However, despite their promise, these techniques exhibit important gaps that limit their effectiveness in the real-world, semi-literate contexts.

The modality of speech is promising for accessibility, however for communication across communities with significant linguistic diversity (no common standard spoken language), sending voice messages does not help. For instance, in India, where languages and dialects change every few hundred kilometers, voice-based communication cannot reliably support broader accessibility or facilitate interaction across linguistically diverse communities. 

Similarly VLMs can possibly generate intermediate images, but it has been observed in earlier \enquote{text to image} systems that the choice of the icon for a particular situation/entity is not consistent and the number of icons and their interplay is also non deterministic. VLMs typically produce monolithic visual outputs rather than structured representations. This lack of hierarchical organization limits opportunities for contextual disambiguation and, consequently, hampers the comprehensibility of the final output.

\section{Related Work}

\subsection{Visual forms of communication}
Visual forms of communication encompassing both linguistic and computational approaches include a variety of forms  including, (i) \textit{Semantographics}, having a fixed set of ideographs to communicate high level semantic ideas \cite{semantographics} (ii) \textit{Emojis and emoticons}, convey emotions, tone and context \cite{emoji} (iii) \textit{Way-finding boards}, help navigate at public places including airports, etc. with direction cues \cite{wayfinding} (iv) \textit{Infographics}, in forms of charts and graphs conveying insights and relationships in data \cite{infographics} (v) \textit{Maps}, for geographical relationships \cite{maps} and (vi) \textit{Gestures and body language}, for communication of emotion and intentions \cite{gestures}. All these visual forms of communication play a crucial role in human interaction, facilitating effective transmission of information across and work across language barriers. 

To the best of our knowledge, none of the existing methods of visual communication incorporate hierarchical simplification on semantic complexity keeping the ideographic inventory smaller in number, leverage the interactive capabilities of digital platforms and are not universal for use. 

\subsection{Neuro-symbolic AI}
Neuro-symbolic AI refers to AI systems that seek to integrate neural network-based methods with symbolic knowledge based approaches \cite{neurosymbolic2,neurosymbolic1,neurosymbolic3}. Embodying intelligent behavior in an AI system must involve both cognition using guiding principles (coming via Symbolic Structures) along with generative and pattern processing capabilities (coming from Neural Structures). Combining both systems enables verifiable reasoning, planning, generalization and abstraction. We use Natural Semantic Metalanguage (NSM) for building Symbolic principles and Generative AI for scale and generation of new concepts.

Natural Semantic Metalanguage (NSM) advocates that all languages in the world irrespective of their origin and timeline contain a semantic core of universals (referred to as semantic primes and molecules in NSM) and this core forms the backbone of human communication \cite{nsm1, nsm2, anna}.  

Generative AI in the recent years, have become the backbone for most NLP tasks primarily text generation/ understanding/ inferencing \cite{entailment}. LLMs encapsulate world's knowledge in form of parametric memory and can be very instrumental in referencing new concepts not seen in model trainings \cite{gpt,llama}. 

\section{Problem Discovery}

To the best of our knowledge, no prior work directly aligns with our approach of combining minimal textual elements for syntax with symbolic representations as the primary carriers of semantics. Existing studies largely fall into three categories: purely symbolic systems, text-based approaches, or the use of emoticons and emojis, which are typically employed as paralinguistic cues (e.g., to signal emotions or attitudes). Consequently, there is no directly comparable baseline in the literature, and we therefore adopt a first-principles approach to problem formulation, system design, and solution development.

\subsection{User Needs Assessment and Benchmarking}

\noindent We partner with a mix of urban and rural semi-literate subjects from Indian demography to identify their challenges and expectations from digital communication platforms. A voluntary group of 20 semi-literates (academic education less than twelve years and face challenges with digital communication) participants $(P = \{ p_j | j = 1,2,...20\})$ were engaged, to discover problems, preferences, and challenges. We illustrate further details about the users engaged in Appendix \ref{appendix:B}.

We request the participants to use existing ideographic methods (BETA, SCLERA and ARASAAC) \cite{scalera, arasaac} and interpret a set of twenty five sample messages $(M = \{ m_i | i = 1,2,...25\})$ over a period of five days (which means five ideographic messages each day). The participants were made familiar with these scripts (not an exhaustive training though), at the start of the workshop and feedback provided on each day based on how they responded. Each day, we collect interpretations of the participants (as text) for each message $m_i^j$ (\textit{i} is message id and \textit{j} is participant id) and compare these with the actual message text ($a_i$) used to generate the initial message $m_i$. We use comprehensibility (to measure usability) and learnability (to measure learning potential) as essential metrics for benchmarking current methods, and we use the same method (plus additional metrics on user experience, and effectiveness) later when we validate our method. The results are illustrated in Table \ref{tab:sota}, and further in Appendix \ref{appendix:A}.

Comprehensibility measures how easily users can understand the language, which can be measured by evaluating semantic comprehension of sample messages. using Meteor $(M)$ as a metric for comprehensibility, which compares common words considering synonyms, stemming, and word order \cite{meteor}. For each participant $p_j$, for each day we compare the message interpretations with actual message text $\{ \frac{1}{5} M (m_i^j, a_i) | i = 1,2,...5\}$ and compute average comprehensibility for each participant $(c_j)$. Finally, we average the scores for all 20 participants and calculate the final comprehensibility $(c)$ for each day. 

Learnability measures how quickly users can learn to use the language proficiently. We also consider the phenomenon of \textit{Plateau Effect} in learning, which occurs because initial gains in proficiency are typically easier and faster compared to mastering finer details or nuances \cite{learnabilityMetric}. We use normalized learnability metric calculated as the mean of normalized difference of comprehensibility on day 5 $(c_{5})$ and day 1 $(c_{1})$ on comprehensibility along with penalty to accommodate plateau effect $({(c_{5} -c_{1})}/{c_{1}}) \dot W_t$ where $W_t = {1-(\left|c_5-T\right|} / {T}) $, and T is the learning efficiency threshold, at which the rate of learning slows down \cite{learning} (we use 90\% as threshold value). 

\begin{table}[t]
\centering
\begin{tabular}{p{4.2em}  c c}
\hline
 &  \textbf{Comprehensibility} &  \textbf{Learnability}\\
 \hline 
 \textbf{SCLERA}  &  {0.42 - 0.46} &  {0.049}\\
 \textbf{BETA} &  {0.38 - 0.41} &  {0.036}\\
 \textbf{ARASAAC} &  {0.43 - 0.46} &  {0.036}\\ 
\hline
\end{tabular}
\caption{Benchmarking existing pictographic methods. (Values for comprehensibility are day (1 - 5) numbers)}
\label{tab:sota}
\centering
\end{table}

\subsection{Dataset Identification } 

While user assessment helps to identify existing issues and benchmark current methods, it is very important to identify an appropriate and relevant dataset which can provide meaningful insights in the semi-literate texting behavior enabling better generalization to real-world scenario. We use an existing dataset (3000+ messages; 300+ respondents) from rural and urban Indian demographics, manually collected by talking to semi-literates \cite{dataset}. 

\subsection{Defining the Ideographic Scope} 

In order to achieve high comprehensibility and learnability, it is important to balance semantic simplification (for complex words) and syntactic preservation (keeping some text as is). We partition the words in the sentence into two categories, the first category contains complex words that needs to be simplified and ideographed, while the second category comprises easy to understand text that acts as the binding element, maintaining sentence structure and flow.

It has been observed that readability scores give an indication of the ease of comprehension primarily on lexical complexity, familiarity, legibility and typography. We use a variety of readability scores to evaluate the complexity of various words in the identified dataset \cite{flesch,dale,gunning,kincaid} and conclude that nouns are most complex concepts in the sentence, followed by verbs. We also observe that majority of the dataset volume (66\% in this corpus) is composed of nouns and verbs. It is also noted that the length of text message exchanged is limited to 10 words at the most (Mean 7.2, Median 7, Mode 6). We also cross validate our observations on another corpus (Sketch Engine containing a corpus of 36 billion words) and confirm similar findings \cite{sketch1,sketch2}. Hence, we scope our work for small short text messages ($\leq$20 words) along with enabling semantic simplification for all nouns and complex verbs (preserving the rest of text as binding element).

\section{System Design}

As illustrated in Figure \ref{summary}, we use Neuro-symbolic AI architecture having two separate components for defining guiding principles (Symbolic Structure) and pattern processing abilities (Neural Structure). 

\subsection{Design of Symbolic Structure}
\noindent We overlay the architecture of our symbolic structure on the linguistic theory of NSM so as to establish the hierarchy of foundational semantic concepts (referred as Semantic Universals from here on-wards). 

NSM recommends semantic universals to follow a three level hierarchy, (i) The first level designates semantic concepts in the same broader category referred to as \textit{Semantic Classes (SC)} and consists of categories like human class, event class, etc.; (ii) The second level indicates more specific patterns or finer distinctions within SC and can be described with same constituents. This is referred as \textit{Semantic Templates (ST)}, including concepts like  human relationships etc.; and (iii) The third level consists of elementary concepts as (key, value) pairs referred to as \textit{Semantic Variable (sv) and semantic Molecule (sm)} tuples. Concepts within each ST can be described by same (sv, st) tuples set.

In order to represent these guiding principles in a closed form, we formulate a \textit{Mathematical Foundational Model} for NSM. 

\vspace{3pt}
\noindent Consider

\vspace{5pt}
\noindent \textit{Semantic Class (SC)} = \{$sc_i$ | \textit{i} $\in$ [1,n1]\}

\vspace{1pt}
\noindent \textit{Semantic Template (ST)} = \{$st_i$ | $i$ $\in$ [1,n2]\}

\vspace{1pt}
\noindent \textit{Relationship $(R^{sc/st})$} = $x,\{Y\}$ | $x$ $\in$ $SC$, $Y$ $\subset$ $ST$

\vspace{5pt}
Concepts within each ST, can be represented with same elementary concepts. The entailment of ST into a set of SV (as keys) and SM (as values) tuples is performed manually for the initial set. 

\vspace{5pt}
\noindent \textit{Semantic Variable (SV)} = \{$sv_i$ | $i$ $\in$ $[1,n3]$\}

\vspace{1pt}
\noindent \textit{Semantic Molecule (SM)}  = \{$sm_i$ | $i$ $\in$ $[1,n4]$\}

\vspace{1pt}
\noindent \textit{Entailment (E)} = \{($e_i$ | $i$ $\in$ $[1,n5]$\}

\vspace{1pt}
\noindent \textit{where}  $e_i$ = \{($sv_i, sm_j$) | $i$ $\in$ $[1,n3]$ \& $j$ $\in$ $[1,n4]$\}

\vspace{1pt}
\noindent $\forall$  $sv_i$ $\in$ $SV$ \& $\forall$ $sm_j$ $\subset$ $SM$

\vspace{1pt}
\noindent \textit{Relationship} ($R^{st/e}$) = $x,\{y\}$ | $x$ $\in$ $ST$ \& $y$ $\in$ $E$

\vspace{5pt}
Since the scope of ideographic conversion in our work is limited to nouns and complex verbs (which excludes linking verbs like \textit{'is'}, \textit{'am'} etc.), we build an end-to-end pipeline and tokenize the sentence into its components and conduct parts of speech (POS) classification to simplify them into semantic universals as follows.

\vspace{5pt}
\noindent \textit{Sentence (S)} : \{\:$w_i^{all}$ | $i$ $\in$ $[1,n]$ \}

\vspace{1pt}
\noindent 
\textit{Ideographic Words (IW)} : \{$w_j^{id}$ | $j$ $\in$ $[1,n']$\}

\vspace{1pt}
\noindent \textit {IW} $\subseteq$ $S$ \space ; \noindent $\forall$ $POS(w_j^{id})$ $\in$ \{$Noun, Verb$\}

\vspace{1pt}
\noindent $\forall$ ($w_j^{id}$) $\in$ ($sc_k,st_l$) 

\vspace{1pt}
\noindent $\forall$ ($st_{l}$) $\in$ \{($sv_1,sm_1$), ($sv_2,sm_2$),...,($sv_{m},sm_{m}$)\} 

\vspace{5pt}
We illustrate some examples, to further clarify the process of semantic simplification within \textit{Humans} (SC) and \textit{Human Relationships} (ST).

\vspace{5pt}
\noindent
\noindent $Mother$ $\equiv$ $(Path,P) (Gender,F)$ 

\vspace{1pt}
\noindent $Nephew$ $\equiv$ $(Path,S_i, C) (Gender,M)$

\vspace{1pt}
\noindent $Grandfather$ $\equiv$ $(Path,P; P) (Gender,M)$ 

\vspace{1pt}
\noindent $\textit{(M-Male, F-Female, P-Parent, $S_i$-Sibling, C-Child)}$

\vspace{5pt}
Having the mathematical foundational model in place, the next steps is to build an initial ontology for concepts (nouns and verbs, present in our initial dataset) into hierarchical semantic structure.

\vspace{25pt}
\noindent \textbf{Building Initial Ontology}
\vspace{2pt}

\noindent It has been observed that \textit{Semantic Relatedness (SR)} can be most appropriately evaluated via dense vector representations \cite{semanticSurvey}. We have evaluated shallow, context independent distributed vector representations Word2Vec and FastText along with deep transformer based context dependent embedding BERT and ELMO \cite{W2V, bert, elmo}. We use these embeddings to group together semantic clusters, applying a flat (non hierarchical) clustering algorithms (centroid based and density based) along with hierarchical algorithms (agglomerative and graph based) \cite{birch,cluster1,cluster2}.

\setlength{\textfloatsep}{5pt}
\begin{table}[b]
\centering
\begin{tabular}{p{5.4em}  p{2.2em}  p{2.2em}  p{2.4em}  p{2.4em}}
\hline
 &  \textbf{W2V} &  \textbf{FT} &  \textbf{ELMO} &  \textbf{BERT}\\
 \hline
 \textbf{K-Means}  & \small{0.78} & \small{0.84} &  \small{0.89} & \small{0.92}\\
 \textbf{DBSCAN} & \small{0.79} & \small{0.78} &  \small{0.90} & \small{0.91}\\
 \textbf{BIRCH} & \small{0.81} & \small{0.79} &  \small{0.93} & \textbf{\underline{0.94}}\\
 \textbf{Agglomerative} & \small{0.88} & \small{0.89} & \small{0.72} & \small{0.93}\\
\hline
\end{tabular}

\caption{Semantic clustering approaches and findings. Human evaluation is normalised on 0-1 scale with 1 being best score. (W2V - Word2Vec, FT- FastText)}
\label{tab:cluster}
\centering
\end{table}

As illustrated in Table \ref{tab:cluster}, we use combinations of vector representations and two-level clustering techniques to discover semantic clusters in our initial dataset. We initially find semantic classes (SC) as first-level semantic concepts, and then within each SC we explore semantic templates as next-level semantic structures. We evaluate the results by human validation and observe that using a combination of BERT (for embedding) and BIRCH (as clustering approach) give the most appropriate results. 

Further breakdown of semantic templates into pairs of semantic variables and molecules is accomplished by careful linguistic analysis of each concept manually. We have collaborated with the domain experts for this work, with inputs from our end users. (Appendix \ref{appendix:C} for details) 

\subsection{Design of Neural Structure}

While our initial ontology contains approximately 1550 concepts (1100 noun entities and 450 verb entities), we design a neural method to systematically entail the OOV concepts into the prescribed mathematical model to ensure that our model is universal in use. 

The recent evolution of conversational Artificial Intelligence (AI) specifically LLMs, it is observed that LLMs become very instrumental in reasoning and inferencing use cases. LLMs require natural language stimuli referred to as prompt as input to pass instructions to LLM to get respective output \cite{promptengineering, promptprogramming, prompting}. Ideally the prompt (P) should contain a Context (C) such as background information of the task, the instructions (I) such as directions, constraints and process to follow and examples (E) as the few shot learning methodology to teach the objective of the exercise $(Prompt (P) \equiv LLM\:(C, I, E))$.

The context is the most crucial component in the prompt, to achieve appropriate output from LLM encapsulating intent along with the environment in consideration. We supply the context in the form of conversational sequence using in-context learning method of Tree Of Thoughts (ToT) \cite{tot}. Instructions are provided in forms of rules and deterministic process which needs to be followed, along with the format (and considerations) of output. The examples provided within the prompt strengthens the inferencing behavior. (Appendix \ref{appendix:E} for details)

\subsection{Design for User Experience}

The final purpose of our work is to assist digital enablement of semi-literates, using intuitive user interface hence UI design becomes very critical for the success of our work. We focus on simplicity of interface elements including input controls (enabling local language support as binding text) and navigational components (including clickable features to illustrate hierarchy). 

\begin{figure}
\includegraphics[width=0.46\textwidth]{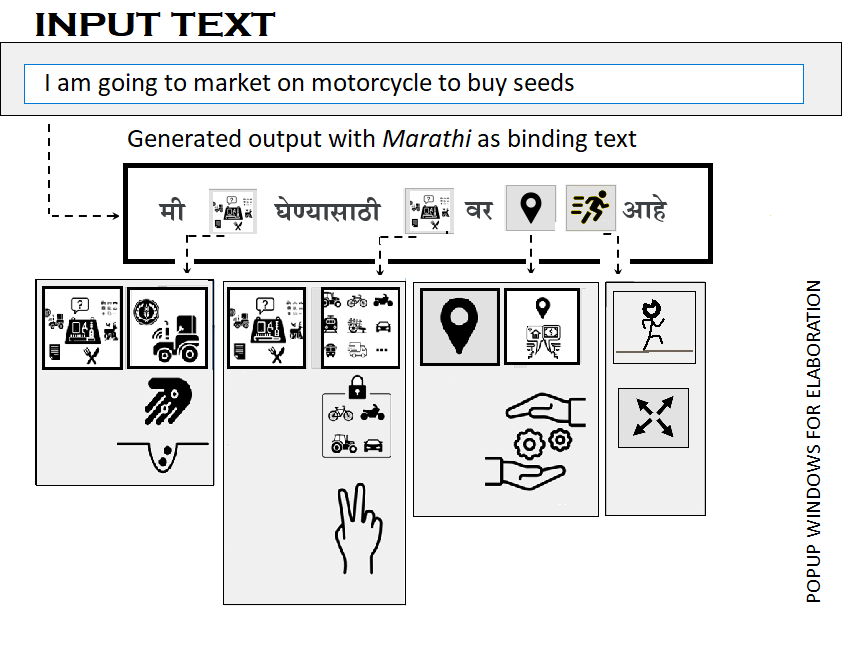}

\caption{Sample output (binding text in Marathi).}
\label{tab:msgs1}
\end{figure}

We refer to \textit{The Noun Project (NP)}\footnote{https://thenounproject.com} for the identification/design of ideographs for our work. NP is a crowd-sourced collection of 3 million icons created by designers from 120+ countries. Each element within semantic class, semantic template and semantic molecule maps to an unique ideograph. Semantic variables are not displayed and kept implicit (based on user feedback). For ideographic selection, we take feedback from our participants to help with the choice of most appropriate icon based on their social and cultural ecosystem (further illustrated in Appendix \ref{appendix:B}). 

\section{Implementation}
We conduct our experiment using Python 3.10.1, PyQt 5 5.15.1, BERT, GPT 3.5 Turbo 16k and Langchain \cite{bert, langchain}. 

\subsection{Pre-Processing}
We perform text cleaning involving removing irrelevant or unwanted elements from the text, such as special characters, punctuation, etc. followed by lower casing, isolating numbers, Part-of-Speech Tagging and lemmatization. We use \textit{nltk package} for most pre-processing tasks which leverages a predefined set of grammatical rules, a dictionary of words/ punctuation/ identifiers, and for the POS tags, it uses Penn Treebank POS tag set. We perform complexity analysis (using readability scores as described in methodology) on nouns and verbs and identify complex words to be ideographed in downstream processes.

\begin{figure}
\includegraphics[width=0.46\textwidth]{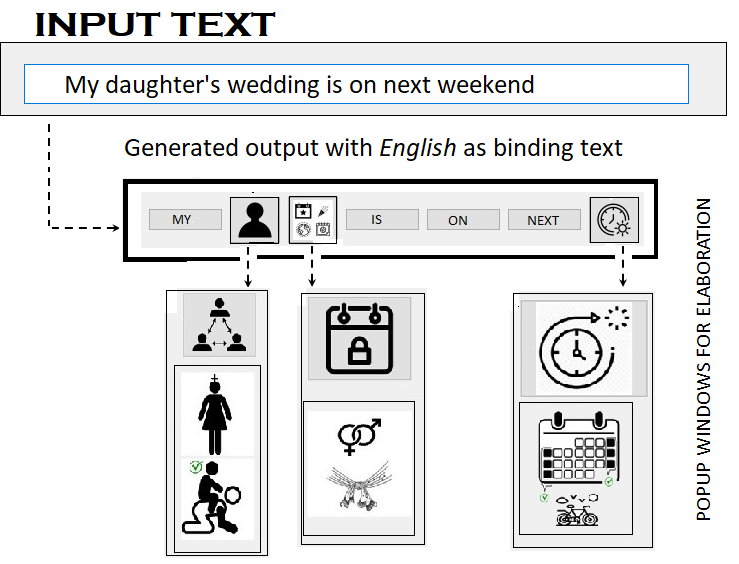}
\caption{Sample output (binding text in English).}
\label{tab:msgs2}
\end{figure}

\subsection{Semantic Decomposition}
Each complex token identified is simplified by breaking up of complex concepts into elementary concepts (SC, ST and (SV, SM) tuples). As described in the methodology, we leverage the initial ontology to identify the tokens and use the pre-built hierarchical semantic decomposition for these concepts.

For concepts not in initial ontology (which we observe only for rare scenarios, within the same user group) we leverage the power of Gen-AI to entail the concepts into elementary concepts as already illustrated in earlier sections. 

Benchmarking results indicate that GPT surpasses other LLMs in inference tasks \cite{legalllm,gptunderstands}. Once the new concepts decompose into hierarchical semantic schema as required, we also add the same into our ontology for future reference. The binding text is translated and rearranged to native language so that the final message uses the semantics and syntax of end user's native language. We use off the shelf \enquote{Google Translate} (via \textit{googletrans} library in python) for translation to native languages, hence the support for native languages is restricted as supported by Google Translate. (See Appendix \ref{appendix:F} for details)

\subsection{Presentation Framework}
NIM is displayed as a sequence of ideographic semantic classes ($sc_i$) with binding text in the initial view. Semantic decomposition into semantic template and further into semantic variable, semantic molecule pairs appear when the corresponding $sc$ icon is clicked. The purpose of keeping semantic details ($st, sm, sv$) information clickable is to keep the final multimodal sentence in a single view (to avoid distraction by not passing too much of information together) and display hierarchical breakdown of information as and when requested.  

Figure \ref{tab:msgs1} and \ref{tab:msgs2} illustrates sample output from our system using Marathi and English as binding text. The syntax of sentence (and hence the location of ideographs) changes with the choice of binding text. The boxes marked with incoming arrows, appear as popups when the corresponding icons are clicked. 

\section{Results and Analysis}
To ensure a robust real-world validation of NIM, we collaborate with a separate cohort of 200 participants, distinct from those involved in the design process to minimize bias and provided compensation for their participation. The testing procedure spawned over a period of one week where each day the participants interpret a set of sentences, and we survey their response on multiple parameters. 

There is actually no similar work which combines bare minimum elementary text (for syntax) with symbols doing heavy lifting for semantics. Most existing work are either all symbols or text or emoticons (for emotional emphasis). Some latest VLM work is on the other side very non deterministic and does not contain to the boundaries of culture and semantic hierarchy. 

Our evaluation approach broadly encompasses three dimensions: \textbf{cognitive accessibility and user engagement}, assessing comprehensibility, learnability, and user appeal; \textbf{semantic robustness}, evaluating ideograph effectiveness, generalization to out of vocabulary concepts, and the adequacy of concept inventory; \textbf{design rationale}, using ablation studies to validate the contribution of individual components. These dimensions provide critical insights into the system's capabilities, detailed in the following sections.

\subsection{Comprehensibility and Learnability}
We use the method illustrated in \S 3.1 for empirically validating semantic comprehensibility and learnability, but with extended participants and metrics.  

We consider 50 sentences stochastically selected from the dataset \cite{dataset} and translate them to NIM. We distribute these transformed sentences into 5 sets (10 sentences each) and share each set every day for a period of 5 days, with our participants. We ask the participants to share their interpretation every day (verbally or written), and candid feedback is given to them on their response. Not to mention that that every participant is also run through a basic enablement session on NIM before they start.

We use statistical harmonic mean metric METEOR \cite{meteor}, along with sentence transformers (S-BERT \cite{sbert}, MPNet \cite{mpnet} and all-MiniLM-L6 \cite{miniLM}) for semantic textual similarity (STS) of human interpretation (with baseline sentences). For learnability, we use normalized learning rate with reward for reaching the threshold faster, reflecting their potential for practical use and impact.

Table \ref{tab:humanevaluation} summarizes the results (mean values) on all metrics over a period of 5 days and Table \ref{tab:learningCurve} illustrates the learning rates which is the differential comprehensibility rate over 5 days period (explained in \S 3.1). 

The empirical results as illustrated in Table \ref{tab:humanevaluation} and Table \ref{tab:learningCurve} indicate that, comprehensibility improves with time. We also observe that NIM reaches a decent value of more than 80\% by day 5, reflecting good learning rates for our end users. Comparing with benchmarking studies we observe an improvement of 1.9 X in comprehensibility and 8.5 X in learnability. While it may not be appropriate to directly compare with SOTA due to variety of end users, and purpose of other systems, we have shown these results to illustrate the value of NIM in semi-literate context. 

\begin{table}
\begin{tabular}{p{0.35cm} p{0.42cm} p{0.42cm} p{0.42cm} p{0.42cm} p{0.42cm} p{0.42cm} p{0.42cm} p{0.42cm}}
\toprule

\multirow{2}{*}{\textbf{D}} & \multicolumn{2}{c}{\textbf{Meteor}}  &  \multicolumn{2}{c}{\textbf{SBERT}}  &  \multicolumn{2}{c}{\textbf{MPNet}}  &  \multicolumn{2}{c}{\textbf{MiniLM}}  \\
 & $\mu$ & $\sigma$ &  $\mu$ & $\sigma$ & $\mu$ & $\sigma$ & $\mu$ & $\sigma$ \\

\hline
{\#1} & \small{0.57} & \small{0.02} & \small{0.62} & \small{0.03} & \small{0.62} & \small{0.03} & \small{0.63} & \small{0.03}\\
{\#2} & \small{0.62} & \small{0.03} & \small{0.67} & \small{0.01} & \small{0.67} & \small{0.01} & \small{0.72} & \small{0.04}\\
{\#3} & \small{0.65} & \small{0.03} & \small{0.74} & \small{0.03} & \small{0.71} & \small{0.03} & \small{0.74} & \small{0.03}\\
{\#4} & \small{0.72} & \small{0.03} & \small{0.84} & \small{0.02} & \small{0.81} & \small{0.03} & \small{0.81} & \small{0.03}\\
{\#5} & \small{0.81} & \small{0.03} & \small{0.84} & \small{0.01} & \small{0.86} & \small{0.04} & \small{0.82} & \small{0.05}\\
\hline
\bottomrule
\end{tabular}
\vspace{-5pt}
\caption{Empirical Evaluation (0-1 scale) on Meteor, S-BERT, MPNet and miniLM over a period of 5 days(D).}

\label{tab:humanevaluation}
\end{table}

\begin{table}[b]
\begin{tabular}{p{1.1cm} p{1.1cm} p{1.1cm} p{1.1cm} p{1.1cm}}
\toprule
\hline
 & {\textbf{Meteor}} &  {\textbf{SBERT}} &  {\textbf{MPNet}} &  {\textbf{MiniLM}} \\
\hline
{\textbf{LCR} $\uparrow$} & {0.381} & {0.331} & {0.369} & {0.273}\\
\hline
\bottomrule
\end{tabular}

\caption{Learning Curve Rate (LCR) evaluation on various learning metric (similar to Table \ref{tab:sota})}
\label{tab:learningCurve}
\end{table}

\begin{table}
\begin{tabular}{p{0.35cm}  p{0.42cm} p{0.42cm} p{0.42cm} p{0.42cm} p{0.42cm} p{0.42cm} p{0.42cm} p{0.42cm}}
\toprule
\multirow{2}{*}{\textbf{{D}}} & \multicolumn{2}{c}{\textbf{{Exp}}}  &  \multicolumn{2}{c}{\textbf{{EOU}}}  &  \multicolumn{2}{c}{\textbf{{ITR}}}  &  \multicolumn{2}{c}{\textbf{{Int}}}  \\
 & $\mu$ & $\sigma$ &  $\mu$ & $\sigma$ & $\mu$ & $\sigma$ & $\mu$ & $\sigma$ \\

\hline
{\#1} & \small{6.30} & \small{1.56} & \small{6.22} & \small{1.40} & \small{6.36} & \small{1.32} & \small{6.35} & \small{2.10}\\
{\#2} & \small{6.35} & \small{1.29} & \small{6.42} & \small{1.32} & \small{6.43} & \small{1.26} & \small{6.79} & \small{2.35}\\
{\#3} & \small{7.67} & \small{1.28} & \small{7.63} & \small{1.30} & \small{7.69} & \small{1.30} & \small{7.48} & \small{1.42}\\
{\#4} & \small{8.25} & \small{1.48} & \small{8.08} & \small{1.19} & \small{8.08} & \small{1.57} & \small{8.36} & \small{1.45}\\
{\#5} & \small{8.21} & \small{1.45} & \small{8.35} & \small{1.45} & \small{8.37} & \small{1.43} & \small{8.40} & \small{1.38}\\
\hline
\bottomrule
\end{tabular}
\vspace{-5pt}
\caption{Empirical evaluation (0-10 scale) on Expressiveness (Exp), Ease of Use (EOU), Intention to re-use (ITR) and being of interest (Int) over a period of 5 days.}
\label{quali2}
\end{table}

\subsection{User Satisfaction}

We have already discovered that the biggest challenge for semi-literates not to use digital communication methods is the lack of engaging content along with non-intuitive user experience. We conduct a qualitative survey each day on 4 parameters namely (i) Expressiveness: ability to conceptualize the semantic characteristics of the inherent concept, (ii) User Experience (UX): ease and convenience for users to interact and understand, (iii) Intention to Reuse: measure of user satisfaction and perceived usefulness and (iv) Interest: measure the value, relevance and appeal to the audience. As illustrated in Table \ref{quali2}, the results (mean values) clearly indicate that the engagement shows a consistent improvement across all measured metrics. 

\subsection{Ideograph Effectiveness}

For ideographic effectiveness, we make use of Multiple Index Approach (MIA) as prescribed by European Telecommunications Standard Institute (ETSI) on the principles of CCITT Recommendations \cite{ccitt, mia}. MIA recommends the design of questionnaire on five parameters namely, Valid associations/ Hit-Rate (HR), Invalid associations/ False Alarm Rate (FAR), Missing Associations (MA), Confidence of association/ Subjective Certainty (SC) and Relevance of association/ Subjective Suitability (SS). We conduct test(s) using a questionnaire on the last day of the evaluation when the participants are familiar with the system and can evaluate based on their overall experience over last 5 days. As illustrated in Table \ref{tab:mia} the results clearly indicate that the ideographs used have high degree of certainty, relevance, confidence and lower misinterpretation. 

\begin{table}[b]
\centering
\begin{tabular}{p{3.0em}  p{3.0em}  p{3.0em}  p{3.0em} p{2.0em}}
\hline
 {\textbf{HR} $\uparrow$} &  {\textbf{FAR} $\downarrow$} &  {\textbf{MA} $\downarrow$} &  {\textbf{SC} $\uparrow$} & {\textbf{SS} $\uparrow$}\\
 \hline
{0.89} & {0.07} & {0.05} & {0.86} & {0.83}\\
\hline
\end{tabular}
\caption{Ideographic effectiveness using MIA.}
\label{tab:mia}
\centering
\end{table}

\subsection{Effects of Generative Methods on New Concepts}

While we observe that size and volume of concepts and ideographs follow Zipf’s law with a logarithmic growth trajectory and becomes stable, however our design enables handling OOV concepts using LLMs. Our prompts consist of Context (C) which contains the brief summary of the symbolic structure (design philosophy), instructions (I) to keep the response precise and categorically respond without elaboration, along with a set of Examples (E) applicable for the particular context. We use Tree-Of-Thoughts (TOT) as an inferencing style for prompt design, in order to structures the reasoning process of LLM into a tree-like hierarchy of thoughts, thus incorporating multistep reasoning, creative exploration, and optimization.

We leverage GPT 3.5 and use the LLMs incrementally in three steps, first we prompt find Semantic Class, next we use the response and further extract semantic template followed by semantic decomposition into semantic variables and molecules. At each step the prompt ingredients (C, I and E) are carefully articulated and generalized (iteratively) to get the most optimized response. We validate our method, on a set of 200 concepts (from our initial ontology, 150 nouns and 50 verbs; we don't use these 200 concepts as examples in prompts), and observe an accuracy of 96\% for identification of SC and ST. For identification of SV, SM tuples we achieve 90\% accuracy. We also tried with other LLM models (Gemini, Claude 3.5, Llama) and frameworks like DSPy \cite{dspy}, and most models worked in close range. 

Given that the Natural Semantic Metalanguage (NSM) theory has been empirically validated as a universal framework and that LLMs excel in the semantic decomposition of novel concepts across diverse cultures, languages, domains, and geographies, it is reasonable to conclude that the proposed metalanguage (NIM) can also demonstrate near-universal applicability. 

\begin{table}[t]
\begin{tabular}{c c c c c c c c}
\toprule
\hline
\multicolumn{2}{c}{\textbf{Meteor}}  &  \multicolumn{2}{c}{\textbf{SBERT}}  &  \multicolumn{2}{c}{\textbf{MPNet}}  &  \multicolumn{2}{c}{\textbf{MiniLM}}  \\
$\mu$ & $\sigma$ &  $\mu$ & $\sigma$ & $\mu$ & $\sigma$ & $\mu$ & $\sigma$ \\

\hline
\small{0.62} & \small{0.08} & \small{0.64} & \small{0.07} & \small{0.63} & \small{0.07} & \small{0.62} & \small{0.09}\\
\hline
\bottomrule
\end{tabular}
\vspace{-5pt}
\caption{Ablation study for impact of binding text.}
\label{tab:ablation}
\end{table}

\subsection{Inventory Adequacy}
\begin{figure}
\centering
{\includegraphics[width=0.35\textwidth]{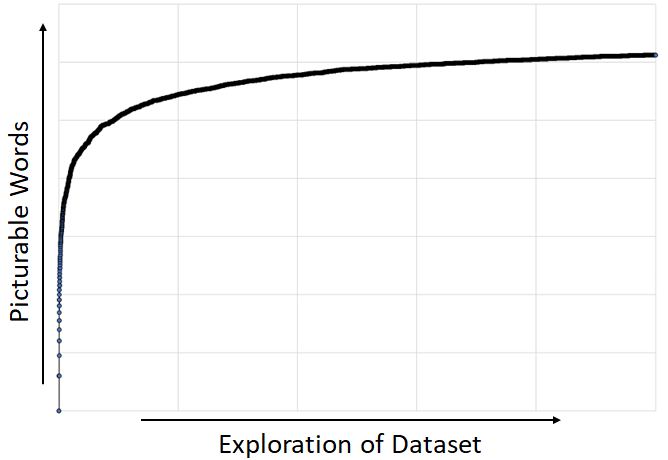}}\\
\vspace{-5pt}
\caption{Growth of picturable words with exploration of dataset.}
\label{volume}
\centering
\end{figure}

Appendix \ref{appendix:D} illustrates the hierchy and size of our ontology to fewer than 500 ideographs to enhance learnability and facilitate efficient comprehension. We also observe that languages often follow power law distribution, i.e., small number of words are used very frequently, while the majority of words are used rarely. We illustrate this phenomena in Figure \ref{volume} using a logarithmic scale on the Y-axis to capture the exponential growth of picturable words during early exploration while highlighting the stabilization phase. 

\subsection{Ablation Studies}
To validate our design choices and evaluate the contributions of multimodal components (combining ideographs with binding text), we conduct ablation study to assess the impact of removing the binding text from the final output, thereby transitioning to a purely ideographic form of communication. These tests were performed on Day 6 of the testing phase with the same participant group. The results, presented in Table \ref{tab:ablation}, reveal a significant drop in comprehensibility (around 24\%) accompanied by an increase in standard deviation (w.r.t. Table \ref{tab:humanevaluation} Day 5 numbers) . This decline is attributed to the universal arrangement of icons, which lacked context from the participants' native languages. Notably, these observations were made after participants had gained familiarity with the script and developed a reasonable level of competency. The findings underscore the critical role of binding text in reducing ambiguity and enhancing comprehensibility.

\section{Conclusion and Future Work}
\enquote{NIM} facilitates digital literacy for semi-literate population across linguistic and geographic boundaries. It embodies the principles of Neuro-symbolic AI combining the generalization strengths of neural networks (LLM's) with structured reasoning from symbolic logic (leveraging the linguistic theory of NSM). Our design of being multimodal, helps with simpler, highly comprehensible, easy to learn, expressive and engaging output on the interactive digital platforms. Our design is extensible for new domains and languages without much incremental effort. 

Our work adopts a participatory, stakeholder-driven design grounded in real-world data from semi-literate users. Using short message datasets and human-in-the-loop evaluation, we co-develop a concept library and communication system that addresses practical digital barriers with contextual ideographs, ensuring both societal relevance and technical robustness.

We envision iterative co-design with semi-literate participants across multiple geographies as a critical direction for future work. Our long-term objective is to open-source this framework, thereby enabling culturally relevant adaptations and fostering broader community-driven innovation. As acknowledged in our Limitations section, a universal set of ideographs may not adequately address the needs of diverse communities; hence, we aim for our methodology to support the creation of localized variants tailored to specific semi-literate or non-literate populations. Furthermore, we plan to develop a stable, production-ready, and robust version of the platform that can be openly shared to encourage collaboration, scalability, and adaptation across diverse regions.

\section*{Limitations}

This study is inevitably shaped by the human centric collaborative methodology with a particular social, cultural, and professional backgrounds of both participants and researchers. This would have inevitably influenced the processes of problem framing, ontology construction, and ideograph selection. We do acknowledge the evolving nature of visual literacy and the risks of misinterpretation. We acknowledge that ideographs carry culturally specific resonances and lose its nuance if designed for cultural neutrality. While our approach sought to mitigate these risks through contextual disambiguation, by rarely showing a symbol in isolation and use compositional layouts and semantic groupings to reinforce meaning and reduce misinterpretation. However we cannot ignore the fact that misinterpretation remains an inherent characteristic of pictographic communication.

It is also important to emphasize that our system is not conceived as a replacement for speech or text, but rather as a complementary modality, designed to augment communication in contexts where conventional forms are inaccessible or inadequate.

\section*{Ethical Considerations}
Our work encompasses human-centered NLP, and we have strongly collaborated with a group of participants in the entire process. The selection of participants was random and we started with a list of voluntary people, from which final group was picked randomly. We have followed the distribution across demography (urban/rural, age and gender) from earlier work which worked on similar problem \cite{dataset}. Every participant was paid for the job (at par with local rates). 

The study adhered to stringent ethical guidelines to ensure participant privacy and data security. Participation was voluntary and participants were made aware of the overall process, along with how the data will be used. Informed consent was obtained in writing from all participants, with no collection, storage, or processing of personally identifiable information (PII) or clinical data. The research complied with the “Guidelines for Ethical Considerations in Social Research \& Evaluation in India” (CMS, 2020), and a self-administered Ethics Sensitivity Test based on these guidelines yielded a “Very Good” grade. Safety measures included conducting all tests in closed-door settings to safeguard participant anonymity.

\bibliography{custom}

\begin{thebibliography}{54}
\providecommand{\natexlab}[1]{#1}

\bibitem[{Bach et~al.(2022)Bach, Sanh, Yong, Webson, Raffel, Nayak, Sharma, Kim, Bari, Fevry et~al.}]{prompting}
Stephen~H Bach, Victor Sanh, Zheng-Xin Yong, Albert Webson, Colin Raffel, Nihal~V Nayak, Abheesht Sharma, Taewoon Kim, M~Saiful Bari, Thibault Fevry, and 1 others. 2022.
\newblock Promptsource: An integrated development environment and repository for natural language prompts.
\newblock \emph{arXiv preprint arXiv:2202.01279}.

\bibitem[{Baker and McCallum(1998)}]{cluster1}
L~Douglas Baker and Andrew~Kachites McCallum. 1998.
\newblock Distributional clustering of words for text classification.
\newblock In \emph{Proceedings of the 21st annual international ACM SIGIR conference on Research and development in information retrieval}, pages 96--103.

\bibitem[{Banerjee and Lavie(2005)}]{meteor}
Satanjeev Banerjee and Alon Lavie. 2005.
\newblock Meteor: An automatic metric for mt evaluation with improved correlation with human judgments.
\newblock In \emph{Proceedings of the acl workshop on intrinsic and extrinsic evaluation measures for machine translation and/or summarization}.

\bibitem[{Bautista et~al.(2017)Bautista, Herv{\'a}s, Hern{\'a}ndez-Gil, Mart{\'\i}nez-D{\'\i}az, Pascua, and Gerv{\'a}s}]{aratraductor}
Susana Bautista, Raquel Herv{\'a}s, Agust{\'\i}n Hern{\'a}ndez-Gil, Carlos Mart{\'\i}nez-D{\'\i}az, Sergio Pascua, and Pablo Gerv{\'a}s. 2017.
\newblock Aratraductor: text to pictogram translation using natural language processing techniques.
\newblock In \emph{Proceedings of the XVIII International Conference on Human Computer Interaction}.

\bibitem[{B{\"o}cker(1996)}]{mia}
Martin B{\"o}cker. 1996.
\newblock A multiple index approach for the evaluation of pictograms and icons.
\newblock \emph{Computer Standards \& Interfaces}.

\bibitem[{Canter(1996)}]{wayfinding}
David~V Canter. 1996.
\newblock Way-finding and signposting: Penance or prosthesis?
\newblock Dartmouth Publishing Company.

\bibitem[{Comito et~al.(2019)Comito, Forestiero, and Pizzuti}]{cluster2}
Carmela Comito, Agostino Forestiero, and Clara Pizzuti. 2019.
\newblock Word embedding based clustering to detect topics in social media.
\newblock In \emph{IEEE/WIC/ACM International Conference on Web Intelligence}, pages 192--199.

\bibitem[{Dada et~al.(2013)Dada, Huguet, and Bornman}]{pcs}
Shakila Dada, Alice Huguet, and Juan Bornman. 2013.
\newblock The iconicity of picture communication symbols for children with english additional language and mild intellectual disability.
\newblock \emph{Augmentative and Alternative Communication}, 29(4):360--373.

\bibitem[{de~Salvo~Braz et~al.(2006)de~Salvo~Braz, Girju, Punyakanok, Roth, and Sammons}]{entailment}
Rodrigo de~Salvo~Braz, Roxana Girju, Vasin Punyakanok, Dan Roth, and Mark Sammons. 2006.
\newblock An inference model for semantic entailment in natural language.
\newblock In \emph{Machine Learning Challenges. Evaluating Predictive Uncertainty, Visual Object Classification, and Recognising Tectual Entailment: First PASCAL Machine Learning Challenges Workshop, MLCW 2005, Southampton, UK, April 11-13, 2005, Revised Selected Papers}, pages 261--286. Springer.

\bibitem[{Deliyannis et~al.(2008)Deliyannis, Simpsiri, and Tsirigoti}]{makaton}
Ioannis Deliyannis, Christina Simpsiri, and Plateia Tsirigoti. 2008.
\newblock Interactive multimedia learning for children with communication difficulties using the makaton method.
\newblock In \emph{International Conference on Information Communication Technologies in Education}, pages 10--12.

\bibitem[{Devlin et~al.(2018)Devlin, Chang, Lee, and Toutanova}]{bert}
Jacob Devlin, Ming-Wei Chang, Kenton Lee, and Kristina Toutanova. 2018.
\newblock Bert: Pre-training of deep bidirectional transformers for language understanding.
\newblock \emph{arXiv preprint arXiv:1810.04805}.

\bibitem[{Face(2022)}]{miniLM}
Hugging Face. 2022.
\newblock \href {https://huggingface.co/sentence-transformers/all-MiniLM-L6-v2} {Sentence-transformers}.

\bibitem[{Farr et~al.(1951)Farr, Jenkins, and Paterson}]{flesch}
James~N Farr, James~J Jenkins, and Donald~G Paterson. 1951.
\newblock Simplification of flesch reading ease formula.
\newblock \emph{Journal of applied psychology}, 35(5):333.

\bibitem[{Goddard(2006)}]{nsm2}
Cliff Goddard. 2006.
\newblock Natural semantic metalanguage.

\bibitem[{Goddard and Goddard(2018)}]{anna}
Cliff Goddard and Goddard. 2018.
\newblock \emph{Minimal English for a global world}.

\bibitem[{Grange and Mulla(2015)}]{learning}
Philippe Grange and Mubashir Mulla. 2015.
\newblock Learning the “learning curve”.
\newblock \emph{Surgery}, 157(1):8--9.

\bibitem[{Grossman et~al.(2009)Grossman, Fitzmaurice, and Attar}]{learnabilityMetric}
Tovi Grossman, George Fitzmaurice, and Ramtin Attar. 2009.
\newblock A survey of software learnability: metrics, methodologies and guidelines.
\newblock In \emph{Proceedings of the sigchi conference on human factors in computing systems}, pages 649--658.

\bibitem[{Guha et~al.(2024)Guha, Nyarko, Ho, R{\'e}, Chilton, Chohlas-Wood, Peters, Waldon, Rockmore, Zambrano et~al.}]{legalllm}
Neel Guha, Julian Nyarko, Daniel Ho, Christopher R{\'e}, Adam Chilton, Alex Chohlas-Wood, Austin Peters, Brandon Waldon, Daniel Rockmore, Diego Zambrano, and 1 others. 2024.
\newblock Legalbench: A collaboratively built benchmark for measuring legal reasoning in large language models.
\newblock \emph{Advances in Neural Information Processing Systems}, 36.

\bibitem[{Gunning(1969)}]{gunning}
Robert Gunning. 1969.
\newblock The fog index after twenty years.
\newblock \emph{Journal of Business Communication}.

\bibitem[{Herv{\'a}s et~al.(2020)Herv{\'a}s, Bautista, M{\'e}ndez, Galv{\'a}n, and Gerv{\'a}s}]{arasaac}
Raquel Herv{\'a}s, Susana Bautista, Gonzalo M{\'e}ndez, Paloma Galv{\'a}n, and Pablo Gerv{\'a}s. 2020.
\newblock Predictive composition of pictogram messages for users with autism.
\newblock \emph{Journal of Ambient Intelligence and Humanized Computing}, 11:5649--5664.

\bibitem[{Hitzler et~al.(2024)Hitzler, Ebrahimi, Sarker, and Stepanova}]{neurosymbolic3}
Pascal Hitzler, Monireh Ebrahimi, Md~Kamruzzaman Sarker, and Daria Stepanova. 2024.
\newblock Neuro-symbolic ai and the semantic web.

\bibitem[{Hitzler and Sarker(2022)}]{neurosymbolic2}
Pascal Hitzler and Md~Kamruzzaman Sarker. 2022.
\newblock Neuro-symbolic artificial intelligence: The state of the art.

\bibitem[{https://www.itu.int/rec/T REC-E.121(1996)}]{ccitt}
https://www.itu.int/rec/T REC-E.121. 1996.
\newblock Ccitt recomendations e.121.

\bibitem[{Imai(2005)}]{gestures}
Gary Imai. 2005.
\newblock Gestures: Body language and nonverbal communication.
\newblock \emph{Retrieved Oct}.

\bibitem[{Khattab et~al.(2023)Khattab, Singhvi, Maheshwari, Zhang, Santhanam, Vardhamanan, Haq, Sharma, Joshi, Moazam et~al.}]{dspy}
Omar Khattab, Arnav Singhvi, Paridhi Maheshwari, Zhiyuan Zhang, Keshav Santhanam, Sri Vardhamanan, Saiful Haq, Ashutosh Sharma, Thomas~T Joshi, Hanna Moazam, and 1 others. 2023.
\newblock Dspy: Compiling declarative language model calls into self-improving pipelines.
\newblock \emph{arXiv preprint arXiv:2310.03714}.

\bibitem[{Kilgarriff et~al.(2004)Kilgarriff, Rychly, Smrz, and Tugwell}]{sketch1}
Adam Kilgarriff, Pavel Rychly, Pavel Smrz, and David Tugwell. 2004.
\newblock Itri-04-08 the sketch engine.
\newblock \emph{Information Technology}, 105(116):105--116.

\bibitem[{Kincaid et~al.(1975)Kincaid, Fishburne~Jr, Rogers, and Chissom}]{kincaid}
J~Peter Kincaid, Robert~P Fishburne~Jr, Richard~L Rogers, and Brad~S Chissom. 1975.
\newblock Derivation of new readability formulas (automated readability index, fog count and flesch reading ease formula) for navy enlisted personnel.

\bibitem[{Kunilovskaya and Koviazina(2017)}]{sketch2}
Maria Kunilovskaya and Marina Koviazina. 2017.
\newblock Sketch engine: A toolbox for linguistic discovery.
\newblock \emph{Jazykovedny Casopis}, 68(3):503.

\bibitem[{Liu and Chilton(2022)}]{promptengineering}
Vivian Liu and Lydia~B Chilton. 2022.
\newblock Design guidelines for prompt engineering text-to-image generative models.
\newblock In \emph{Proceedings of the 2022 CHI Conference on Human Factors in Computing Systems}, pages 1--23.

\bibitem[{Liu et~al.(2023)Liu, Zheng, Du, Ding, Qian, Yang, and Tang}]{gptunderstands}
Xiao Liu, Yanan Zheng, Zhengxiao Du, Ming Ding, Yujie Qian, Zhilin Yang, and Jie Tang. 2023.
\newblock Gpt understands, too.
\newblock \emph{AI Open}.

\bibitem[{Lodding(1983)}]{iconicinterfacing}
Kenneth~N Lodding. 1983.
\newblock Iconic interfacing.
\newblock \emph{IEEE Computer graphics and applications}.

\bibitem[{Maguire(1985)}]{review}
Martin~C Maguire. 1985.
\newblock A review of human factors guidelines and techniques for the design of graphical human-computer interfaces.
\newblock \emph{Computers \& graphics}.

\bibitem[{Mavroudis(2024)}]{langchain}
Vasilios Mavroudis. 2024.
\newblock Langchain.

\bibitem[{Peters et~al.(2018)Peters, Neumann, Zettlemoyer, and Yih}]{elmo}
Matthew~E Peters, Mark Neumann, Luke Zettlemoyer, and Wen-tau Yih. 2018.
\newblock Dissecting contextual word embeddings: Architecture and representation.
\newblock \emph{arXiv preprint arXiv:1808.08949}.

\bibitem[{Radford et~al.(2018)Radford, Narasimhan, Salimans, Sutskever et~al.}]{gpt}
Alec Radford, Karthik Narasimhan, Tim Salimans, Ilya Sutskever, and 1 others. 2018.
\newblock Improving language understanding by generative pre-training.

\bibitem[{Reimers and Gurevych(2019)}]{sbert}
Nils Reimers and Iryna Gurevych. 2019.
\newblock Sentence-bert: Sentence embeddings using siamese bert-networks.
\newblock \emph{arXiv preprint arXiv:1908.10084}.

\bibitem[{Reynolds and McDonell(2021)}]{promptprogramming}
Laria Reynolds and Kyle McDonell. 2021.
\newblock Prompt programming for large language models: Beyond the few-shot paradigm.
\newblock In \emph{Extended Abstracts of the 2021 CHI Conference on Human Factors in Computing Systems}, pages 1--7.

\bibitem[{Seargeant(2019)}]{emoji}
Philip Seargeant. 2019.
\newblock \emph{The Emoji Revolution: How technology is shaping the future of communication}.
\newblock Cambridge University Press.

\bibitem[{Sevens et~al.(2015)Sevens, Vandeghinste, Schuurman, and Van~Eynde}]{scalera}
Leen Sevens, Vincent Vandeghinste, Ineke Schuurman, and Frank Van~Eynde. 2015.
\newblock Natural language generation from pictographs.
\newblock In \emph{Proceedings of the 15th European Workshop on Natural Language Generation (ENLG)}, pages 71--75.

\bibitem[{Sharma et~al.(2023)Sharma, Goyal, and Goyal}]{semantographics}
Prawaal Sharma, Navneet Goyal, and Poonam Goyal. 2023.
\newblock Multimodal semantographic metalanguage (msm): A novel methodology for digital enablement of semi-literates.
\newblock In \emph{Proceedings of the 38th ACM/SIGAPP Symposium on Applied Computing}, pages 844--851.

\bibitem[{Sharma et~al.(2021)Sharma, Goyal, and Vinay}]{dataset}
Prawaal Sharma, Navneet Goyal, and MR~Vinay. 2021.
\newblock Semi-literate texting (slt): Survey based text message dataset from digitally semi-literate users in india.
\newblock \emph{Data in Brief}.

\bibitem[{Sheth et~al.(2023)Sheth, Roy, and Gaur}]{neurosymbolic1}
Amit Sheth, Kaushik Roy, and Manas Gaur. 2023.
\newblock Neurosymbolic artificial intelligence (why, what, and how).
\newblock \emph{IEEE Intelligent Systems}, 38(3):56--62.

\bibitem[{Shirreffs(1992)}]{maps}
WS~Shirreffs. 1992.
\newblock Maps as communication graphics.
\newblock \emph{The Cartographic Journal}, 29(1):35--41.

\bibitem[{Siricharoen(2013)}]{infographics}
Waralak~V Siricharoen. 2013.
\newblock Infographics: the new communication tools in digital age.
\newblock In \emph{The international conference on e-technologies and business on the web (ebw2013)}, volume 169174.

\bibitem[{Song et~al.(2020)Song, Tan, Qin, Lu, and Liu}]{mpnet}
Kaitao Song, Xu~Tan, Tao Qin, Jianfeng Lu, and Tie-Yan Liu. 2020.
\newblock Mpnet: Masked and permuted pre-training for language understanding.
\newblock \emph{Advances in Neural Information Processing Systems}, 33:16857--16867.

\bibitem[{Taieb et~al.(2020)Taieb, Zesch, and Aouicha}]{semanticSurvey}
Mohamed Ali~Hadj Taieb, Torsten Zesch, and Mohamed~Ben Aouicha. 2020.
\newblock A survey of semantic relatedness evaluation datasets and procedures.
\newblock \emph{Artificial Intelligence Review}.

\bibitem[{Tatti(2016)}]{iconji}
Kristen Tatti. 2016.
\newblock New iconji language for the symbol-minded-bizwest.
\newblock \emph{BizWest}.

\bibitem[{Touvron et~al.(2023)Touvron, Lavril, Izacard, Martinet, Lachaux, Lacroix, Rozi{\`e}re, Goyal, Hambro, Azhar et~al.}]{llama}
Hugo Touvron, Thibaut Lavril, Gautier Izacard, Xavier Martinet, Marie-Anne Lachaux, Timoth{\'e}e Lacroix, Baptiste Rozi{\`e}re, Naman Goyal, Eric Hambro, Faisal Azhar, and 1 others. 2023.
\newblock Llama: Open and efficient foundation language models.
\newblock \emph{arXiv preprint arXiv:2302.13971}.

\bibitem[{Van~Dam(1984)}]{van}
Andries Van~Dam. 1984.
\newblock Computer software for graphics.
\newblock \emph{Scientific American}.

\bibitem[{Wierzbicka(1996)}]{nsm1}
Anna Wierzbicka. 1996.
\newblock \emph{Semantics: Primes and universals: Primes and universals}.
\newblock Oxford University Press, UK.

\bibitem[{Williams(1972)}]{dale}
Robert~T Williams. 1972.
\newblock A table for rapid determination of revised dale-chall readability scores.
\newblock \emph{The Reading Teacher}.

\bibitem[{Xue et~al.(2021)Xue, Wang, Zhang, Huang, Li, and Zhu}]{W2V}
Xingsi Xue, Haolin Wang, Jie Zhang, Yikun Huang, Mengting Li, and Hai Zhu. 2021.
\newblock Matching transportation ontologies with word2vec and alignment extraction algorithm.
\newblock \emph{Journal of Advanced Transportation}.

\bibitem[{Yao et~al.(2024)Yao, Yu, Zhao, Shafran, Griffiths, Cao, and Narasimhan}]{tot}
Shunyu Yao, Dian Yu, Jeffrey Zhao, Izhak Shafran, Tom Griffiths, Yuan Cao, and Karthik Narasimhan. 2024.
\newblock Tree of thoughts: Deliberate problem solving with large language models.
\newblock \emph{Advances in Neural Information Processing Systems}, 36.

\bibitem[{Zhang et~al.(1997)Zhang, Ramakrishnan, and Livny}]{birch}
Tian Zhang, Raghu Ramakrishnan, and Miron Livny. 1997.
\newblock Birch: A new data clustering algorithm and its applications.
\newblock \emph{Data Mining and Knowledge Discovery}.

\end{thebibliography}

\appendix
\onecolumn

\section{Other Methods}
Illustrative examples of existing visual methods of communication

\label{appendix:A}
\begin{figure}[h]
\centering
{
[A] \fbox{\includegraphics[width=0.44\textwidth]{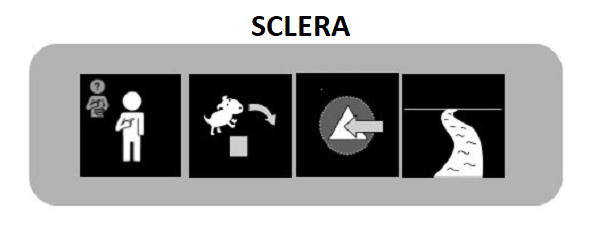}}
[B] \fbox{\includegraphics[width=0.44\textwidth]{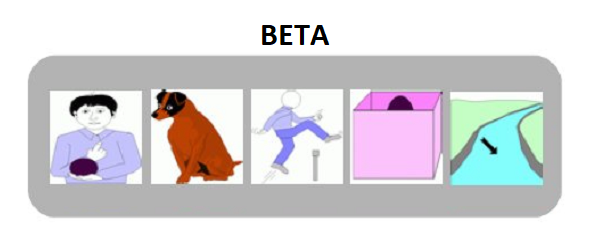}}
}

{
[C] \fbox{\includegraphics[width=0.44\textwidth]{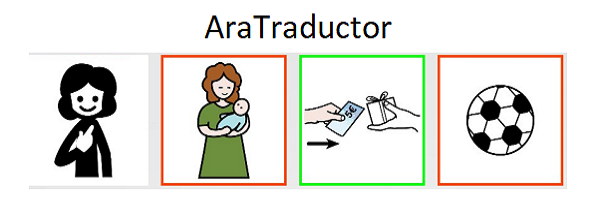}}
[D] \fbox{\includegraphics[width=0.44\textwidth]{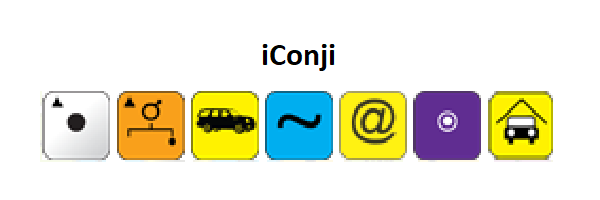}}
}
{
[E] \fbox{\includegraphics[width=0.44\textwidth]{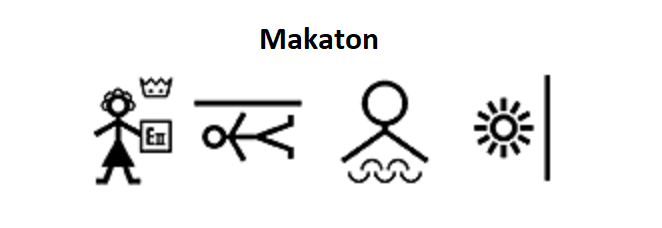}}
[F] \fbox{\includegraphics[width=0.44\textwidth]{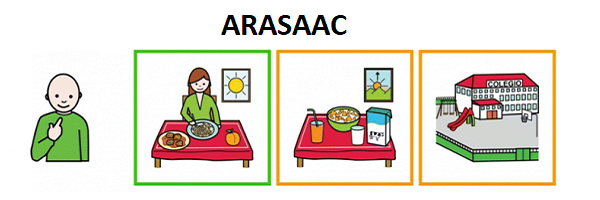}}
}
{
[G] \fbox{\includegraphics[width=0.44\textwidth]{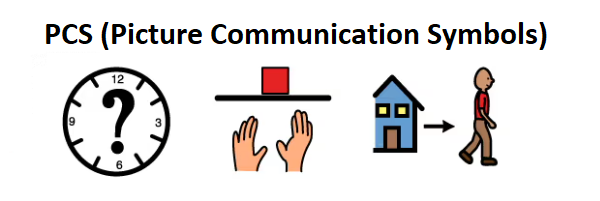}}
[H] \fbox{\includegraphics[width=0.44\textwidth]{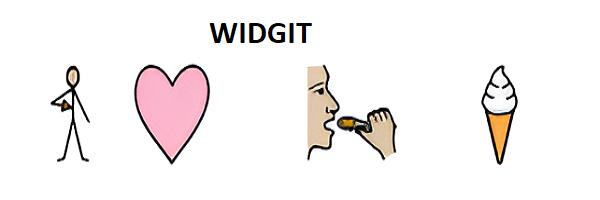}}
}
\end{figure}

\begin{table}[h]
\centering
\begin{tabular}{p{0.5cm} p{2cm} p{6cm} p{5cm}}
\hline
& \textbf{Method} & \textbf{Illustrated Example} & \textbf{Approximate ideograph count}\\
\hline
 A & SCLERA &  My dog jumped in the river. & 3,500 \\
 B & BETA  &   My dog jumped in the river. & 11,500\\
 C & AraTraductor &  My mum bought bought the soccer ball. & 8,500\\
 D & iConji &  My father's SUV is at the garage. & 11,00\\
 E & Makaton &  My queen died peacefully yesterday. & 11,000\\
 F & ARASAAC &  I eat breakfast at school. & 12,000\\
 G & PCS &  When do you want to go. & 37,000\\
 H & Widget &  I like to eat ice cream. & 15,000\\
\hline
\end{tabular}
\end{table}
\vspace{-10pt}
\noindent \cite{scalera, aratraductor, iconji, makaton, arasaac, pcs}

\newpage

\section{Human Engagement}
Metadata (on multiple parameters) for the users engaged during assessment and testing phases.
\label{appendix:B}
\begin{table}[h]
\centering
\begin{tabular}{p{0.5cm} p{3cm} p{5cm} p{4cm}}
\hline
& \textbf{Parameter} & \textbf{Initial assessment} & \textbf{Final assessment}\\
\hline
 1 & Total participants &  20  & 200\\
 2 & Demography &  Rural - 11; Urban - 9  & Rural - 103; Urban - 97\\
 3 & Age group  &   40-50 years & 40-50 years\\
 4 & Gender &  Male - 12; Female - 8 & Male - 123; Female - 77\\
 5 & Duration &  User needs assessment: 1 week & 1 week\\ 
   &   &  Ideographic selection: 1 week & \\ 
\hline
\end{tabular}
\end{table}

\noindent \textbf{Selection criteria} 

\noindent The dataset used in our work \cite{dataset} is a collection of raw data (using door to door survey) for a very similar problem. Our choice for selecting participants on parameters illustrated in the table above is based on this dataset. 

\noindent[Age - Mean 36, Mode - 37, Max 65]

\noindent[Demography - 48\% Urban, 52\% Rural]

\noindent[Gender - Male 58\%, Female 42\%]

\vspace{10pt}
\noindent \textbf{Ideographic selection criteria based on voting (example)} 

\noindent \textit{(Candidate ideographs for representing the concept of \enquote{currency}. The icon in green box is selected as final ideograph based on majority voting.)}
\begin{figure*}[h]
\centering
{
\fbox{\includegraphics[width=0.8\textwidth]{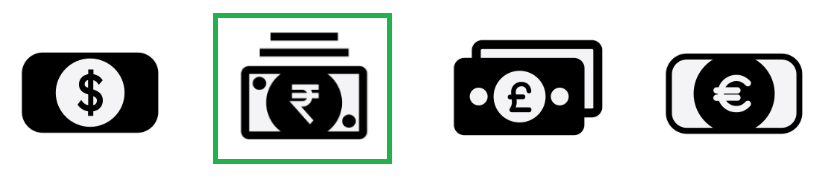}}
}
\end{figure*}

\newpage

\section{Semantic Discovery} 
\textbf{Discovery of Semantic Classes, Semantic Templates, Semantic Variable and Semantic Molecules from the dataset.}
\label{appendix:C}

\noindent \textbf{Step 1:} 
Initially we split the dataset into nouns and verbs via POS analysis and analyse the concepts separately.

\noindent \textbf{Step 2:} For nouns using BERT embeddings and BIRCH clustering approach we find 6 distinctive groups (Semantic Classes) at first level. The image below on the left is an visual representation (reduced dimensions) for the same.

\noindent \textbf{Step 3A:} Within each Semantic Class, we further identify sub clusters which are referred as Semantic Templates in our work. The image on the right is an illustrative example for Semantic Templates within the same Semantic Class.

\begin{figure*}[h]
\centering
{
\fbox{\includegraphics[width=0.45\textwidth]{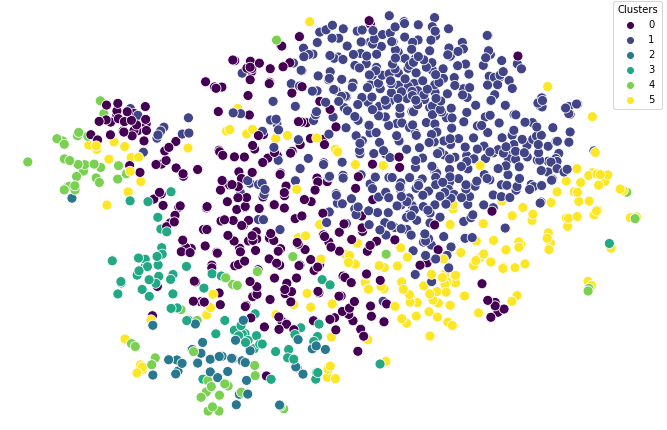}}
\fbox{\includegraphics[width=0.45\textwidth]{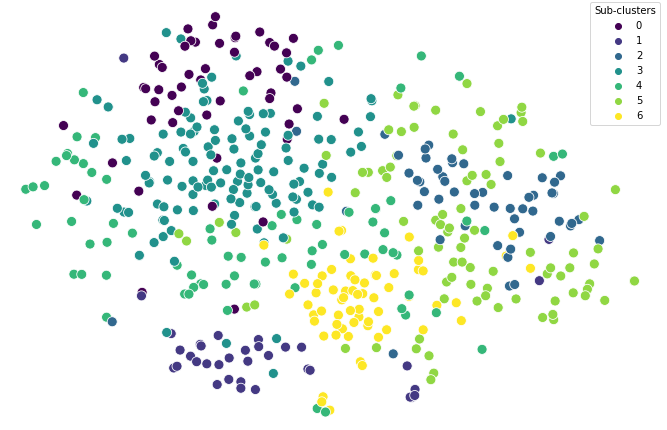}}
}
\caption*{Semantic Classes (Left) and Semantic Templates for Class \#2 (Right) for nouns. Illustrative representation.}
\end{figure*}

\noindent \textbf{Step 3B:} We repeat the same process for Verbs, and identify 3 Semantic classes and further semantic templates for each semantic class. The figure below is a visual representation for verbs. The image on the right is semantic template identification for cluster 2 on the left.

\begin{figure*}[h]
\centering
{
\fbox{\includegraphics[width=0.45\textwidth]{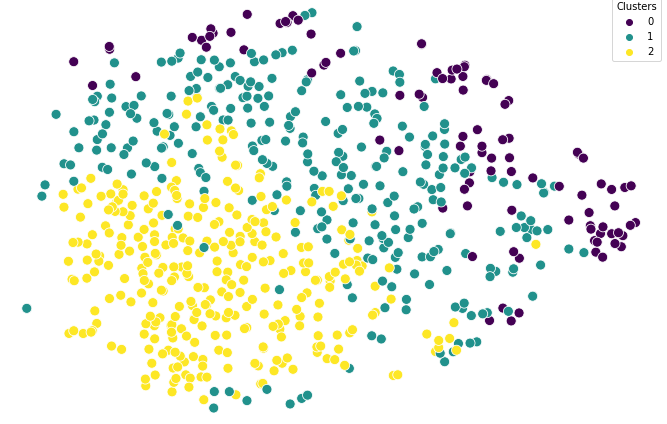}}
\fbox{\includegraphics[width=0.44\textwidth]{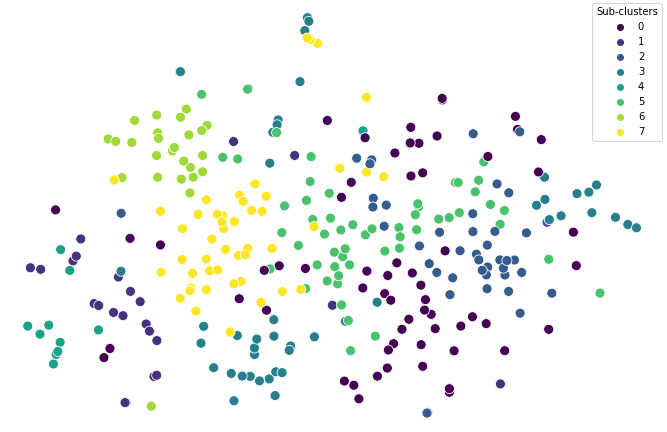}}
}
\caption*{Semantic Classes (Left) and Semantic Templates for Class \#2 (Right) for verbs. Illustrative representation.}
\end{figure*}
\noindent \textbf{Step 4:} Further breakdown into semantic variables and molecules is done manually by careful examination of concepts within the semantic templates. This step is divided into 2 parts (i) Initially the template is generalised/defined as a set of Semantic Variable (SV) and Semantic Molecule (SM) tuples, and (ii) Each concept within the same semantic template is described using the identified (SV, SM) tuple set.

\textit{The entire illustrative count and hierarchy of ontology is illustrated in Appendix D}
\newpage

\section{Ontology}
\label{appendix:D}
\textbf{Hierarchical view}
\begin{figure}[h]
\centering
{
\fbox{\includegraphics[width=0.7\textwidth]{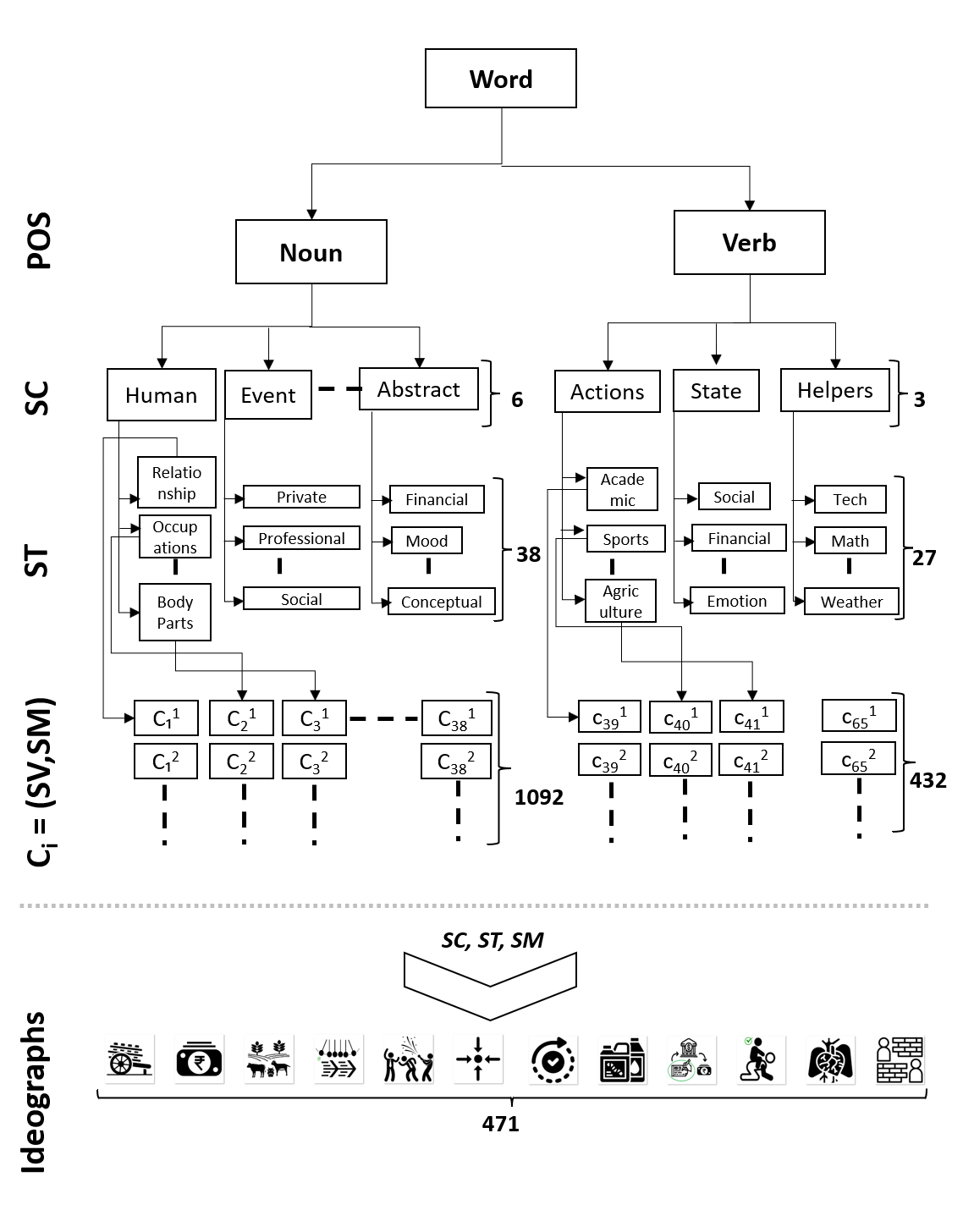}}
}
\end{figure}
\vspace{5pt}

\textbf{Tabular View (summary)}
\begin{table}[h]
\centering
\begin{tabular}{p{3em}  p{2em}  p{2em}  p{3.5em} p{3.5em}}
\hline
  &  {SC} &  {ST} &  {(sm, sv)} & {Ideographs}\\
 \hline
Nouns & {6} & {38} & {1092} & \multirow{2}{*}{{471}}\\
Verbs & {3} & {27} & {432} & \\
\hline
\end{tabular}
\label{tab:size}
\centering
\end{table}

\vspace{5pt}

\small{\textit{$C_i$ issued to represent a unique combination of (SV,SM) tuple set for explication of individual concepts.}}

\newpage

\section{Prompt Engineering}
\textbf{Illustrative examples for OOV concepts using Gen-AI}

\label{appendix:E}
\noindent Consider the following sentence as an illustrative example.

\noindent Input Sentence: \enquote{I am going to mandi on motorcycle to buy seeds}.

\vspace{0.2cm}
\noindent Complex-Words identified - CW1:mandi, CW2:motorcycle, CW3:seeds. 

In the example illustrated, CW2 and CW3 are present in the ontology, while CW1 is Out Of Vocabulary (OOV). Hence for CW2 and CW3, we refer ontology and break them down into semantic universals as below.

\vspace{5pt}

\textit{CW2 - motorcycle} 

\textit{SC : Things}

\textit{ST : Automobile}

\textit{(SV, SM tuples): (Category, Private Transport) (Wheels, Two)}

\vspace{5pt}

\textit{CW3 - seeds}

\textit{SC : Things}

\textit{ST : Agro}

\textit{(SV, SM tuples): (Category, Germinate)}

\vspace{5pt}

For complex words not in ontology (\enquote{Mandi}/CW1 in this example), we use a sequence of prompts using TOT and few-shot learning, where relevant parts of the pre-compiled ontology is passed as context to a LLM and get the recommended response. 

\noindent Input Prompt (P1) - \enquote{Imagine the human annotator has been given the task to hierarchically break down a word into 2 levels. 
Level 1 considers of broad category of word and is referred as SC,
Level 2 considers of narrow category of word and is referred as ST.
Now considering the examples as illustrated \textit{<<Examples with concepts and categorization of SC and ST come here >>}, use your inferencing to find the SC and ST for the word \textit{<<Mandi>>}}

Leveraging response parsing (json parser) and GPT 3.5 as the LLM the final response is illustrated below.

\vspace{5pt}
\noindent Output json = {\enquote{SC}:\enquote{Location},\enquote{ST}:\enquote{Commercial}}
\vspace{5pt}

Once the Semantic Template (ST) is identified (\enquote{Commercial} in this case), we share multiple examples of other concepts in the same ST, within the prompt and leverage LLM to break this down into SV, SM tuples as illustrated below. 

\vspace{5pt}

\noindent Input Prompt (P2) -\enquote{Imagine the human annotator has been given the task to explain the semantic meaning of the word <<Mandi>>, using Key-Value pairs. 
Keys to be considered should be from the predefined set of values \textit{<<add all SV elements for the ST \enquote{Commercial} >>}.
Values considers should be of a predefined set of values \textit{<<add all SM elements for the ST \enquote{Commercial}>>}. Each semantic variable can only take limited values from semantic molecule sets as illustrated \textit{<<For \enquote{Commercial} ST add all semantic variable and molecule combinations here>>}.
Now considering the examples and constraints as illustrated use your inferencing to find the Key-Value pairs for the word \textit{<<Mandi>>}. When there are multiple Key-Value pairs, use Key1,Value1, Key2,Value2 in your response.}

The response received from this prompt appears as:

\noindent Output json = {\enquote{Key1}:\enquote{Purpose},\enquote{Value1}:\enquote{Business}}

Putting all of this together the OOV concept \textit{mandi} is represented in the hierarchical structure as below.

\vspace{5pt}

\textit{CW1 - mandi} 

\textit{SC : Location}

\textit{ST : Commercial}

\textit{(SV, SM) tuples: (Purpose, Business)}

We also tried with other LLM models (Gemini, Claude 3.5, Llama) and frameworks like DSPy \cite{dspy}, and most models worked in close range. Since LLMs are evolving fast we plan to re-validate these models in our future work.

\newpage

\section{Illustrative NIM execution}
\textbf{Step by step process of input text to multimodal output}
\label{appendix:F}
\begin{table*}[h]
\centering  
\begin{tabular}{p{15cm}}
\toprule
\textbf{Text Message (input) in English language for a Marathi end user}:\newline {I am going to market on motorcycle to buy seeds} \\
\vspace{1pt}
\\
\hline
\vspace{8pt}
\textbf{Complex Word identification and Translation (to Marathi)}:\newline {\includegraphics[width=0.8\textwidth, height=8mm]{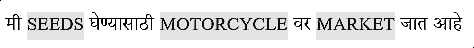}}\\
\hline
\vspace{1pt}
\textbf{Entailment of complex words into hierarchical semantic universals}:\newline \textless elementalization  \textgreater  \newline -\textless cw  \textgreater SEEDS \textless / cw \textgreater \newline --- \textless sc \textgreater things \textless / sc \textgreater  \newline --- \textless st \textgreater agro \textless / st \textgreater \newline ------ \textless sv \textgreater category \textless / sv \textgreater  \textless sm \textgreater germinate \textless / sm \textgreater \newline  - \textless cw  \textgreater MOTORCYCLE \textless / cw \textgreater \newline --- \textless sc \textgreater things \textless / sc \textgreater  \newline --- \textless st \textgreater automobile \textless / st \textgreater \newline ------ \textless sv \textgreater category \textless / sv \textgreater  \textless sm \textgreater private transport \textless / sm \textgreater \newline ------ \textless sv \textgreater wheels \textless / sv \textgreater  \textless sm \textgreater two \textless / sm \textgreater \newline - \textless 
 cw  \textgreater MARKET \textless / cw \textgreater \newline --- \textless sc \textgreater location \textless / sc \textgreater \newline ---   \textless st \textgreater commercial \textless / st \textgreater \newline ------  \textless sv \textgreater purpose \textless / sv \textgreater  \textless sm \textgreater business \textless / sm \textgreater \newline \textless / elementalization  \textgreater\\
\vspace{1pt}
\\
\hline
\vspace{8pt}
\textbf{Final multimodal message output displayed to end user}: \newline\centering{\includegraphics[width=0.8\textwidth, height=60mm]{images/Example1.png}} 
\end{tabular}
\end{table*}

\begin{table*}[t]
\centering
\begin{tabular}{p{15cm}}
\toprule
\textbf{Text Message (input) in English language for a Nepali end user}:\newline {There may be a typhoon tomorrow}\\
\vspace{1pt}
\\
\hline
\vspace{8pt}
\textbf{Complex Word identification and Translation (to Nepali)}:\newline  {\includegraphics[width=0.8\textwidth, height=8mm]{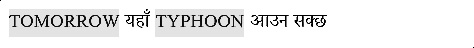}}\\
\hline
\vspace{1pt}
\textbf{Entailment of complex words into hierarchical semantic universals}:\newline
\textless elementalization  \textgreater \newline - \textless cw  \textgreater TOMORROW \textless / cw \textgreater \newline --- \textless sc \textgreater time \textless / sc \textgreater  \newline --- \textless st \textgreater temporal \textless / st \textgreater \newline ------ \textless sv \textgreater marker \textless / sv \textgreater  \textless sm \textgreater day(+1) \textless / sm \textgreater \newline - \textless cw  \textgreater TYPHOON \textless / cw \textgreater \newline --- \textless sc \textgreater event \textless / sc \textgreater  \newline --- \textless st \textgreater event climate \textless / st \textgreater \newline ------ \textless sv \textgreater category \textless / sv \textgreater  \textless sm \textgreater wind \textless / sm \textgreater \newline ------ \textless sv \textgreater intensity \textless / sv \textgreater  \textless sm \textgreater high \textless / sm \textgreater \newline \textless / elementalization  \textgreater
\\
\hline
\vspace{1pt}
\textbf{Final multimodal message output displayed to end user}: \newline\centering{\includegraphics[width=0.8\textwidth, height=60mm]{images/Example2.png}} 
\end{tabular}
\end{table*}

\end{document}